\documentclass[10pt,twocolumn,letterpaper]{article}

\usepackage[pagenumbers]{iccv} 

\definecolor{iccvblue}{rgb}{0.21,0.49,0.74}
\usepackage[pagebackref,breaklinks,colorlinks,allcolors=iccvblue]{hyperref}
\usepackage{algorithm}
\usepackage{algorithmic}
\usepackage{pifont}
\def\paperID{7104} 
\def\confName{ICCV}
\def\confYear{2025}

\title{Unbiased Max-Min Embedding Classification for Transductive Few-Shot Learning: Clustering and Classification Are All You Need}

\author{
Yang~Liu$^{1}$, Feixiang~Liu$^{1}$, Jiale~Du$^{1}$, Xinbo~Gao$^{1,2}$, Jungong~Han$^{3}$ \\
$^{1}$Xidian University, Xi'an, China \\
$^{2}$Chongqing University of Posts and Telecommunications, Chongqing, China \\
$^{3}$Tsinghua University, Beijing, China \\
{\tt\small yangl@xidian.edu.cn, liuf\_x@126.com, 23011211070@stu.xidian.edu.cn, }\\{\tt\small xbgao@mail.xidian.edu.cn, jungonghan77@gmail.com}
}

\begin{document}
\maketitle
\begin{abstract}
Convolutional neural networks and supervised learning have achieved remarkable success in various fields but are limited by the need for large annotated datasets. Few-shot learning (FSL) addresses this limitation by enabling models to generalize from only a few labeled examples. Transductive few-shot learning (TFSL) enhances FSL by leveraging both labeled and unlabeled data, though it faces challenges like the hubness problem. To overcome these limitations, we propose the Unbiased Max-Min Embedding Classification (UMMEC) Method, which addresses the key challenges in few-shot learning through three innovative contributions. First, we introduce a decentralized covariance matrix to mitigate the hubness problem, ensuring a more uniform distribution of embeddings. Second, our method combines local alignment and global uniformity through adaptive weighting and nonlinear transformation, balancing intra-class clustering with inter-class separation. Third, we employ a Variational Sinkhorn Few-Shot Classifier to optimize the distances between samples and class prototypes, enhancing classification accuracy and robustness. These combined innovations allow the UMMEC method to achieve superior performance with minimal labeled data. Our UMMEC method significantly improves classification performance with minimal labeled data, advancing the state-of-the-art in TFSL.

\end{abstract}    
\section{Introduction}
\label{sec:intro}

Convolutional neural networks and supervised learning have propelled significant progress in computer vision, natural language processing, and machine translation, yet they heavily rely on large datasets. In many applications, obtaining annotated data is scarce and expensive \cite{alzubaidi2023survey}, often requiring expert knowledge \cite{spasic2020clinical}, such as for medical image diagnosis. This scarcity hampers the training of effective models.

This limitation has spurred interest in \emph{few-shot learning} (FSL) \cite{parnami2022learning}, which aims to enable models to generalize from only a few labeled examples and thus bridge the gap between data-rich machine learning and the data-scarce realities often encountered in practice \cite{yuan2024spatio}. Despite considerable progress, achieving robust performance under minimal data remains challenging. Approaches such as meta-learning, metric learning, and transfer learning each contribute unique insights to the FSL landscape.

\begin{figure}[t]
  \centering
   \includegraphics[width=0.8\linewidth]{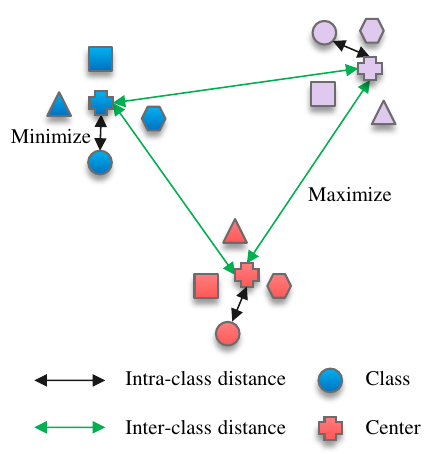}
   \caption{The UMMEC creates a feature space where each class forms a compact cluster, while the clusters of different classes are widely separated.}
   \label{fig1}
\end{figure}

Many recent works \cite{huang2024relation,xi2025transductive,wang2025transductive,zanella2025boosting} focus on transductive few-shot learning (TFSL), which leverages both labeled and unlabeled data from the target task. By incorporating the distribution of unlabeled data, TFSL reduces the data requirement and improves generalization compared to inductive FSL, which relies solely on labeled examples. This approach shows great potential in addressing the inherent challenges of few-shot learning and expanding what can be achieved with limited labeled data.

Despite its promise, TFSL still faces key obstacles. A primary concern is the \emph{hubness} problem \cite{trosten2023hubs}, where certain “hub” points in the embedding space appear disproportionately similar to many others. This leads to an uneven distribution of embeddings that can severely degrade classification performance, causing some classes to be overrepresented while others are underrepresented, ultimately harming generalization.

In TFSL tasks, maintaining a balance between local similarity and global uniformity\cite{guo2022rsnet} is crucial. Existing methods often struggle to preserve local similarities while ensuring that different classes are uniformly distributed in the embedding space. This imbalance can result in poor discrimination between classes, reducing the effectiveness of the model in distinguishing similar but distinct classes.

Another challenge in FSL is optimizing class prototypes and their distances \cite{allen2019infinite}. Traditional methods often use simple averaging, which may fail to capture the complex distribution of samples. Moreover, maintaining well-separated prototypes while keeping samples close to them is vital for high accuracy, yet many approaches lack robust mechanisms to achieve this effectively.

To address these challenges, we propose the Unbiased Max-Min Embedding Classification (UMMEC) method, which comprises two key components: Unbiased Max-Min Embedding (UMME) and Unbiased Max-Min Classification (UMMC). Our approach introduces three main contributions:

\begin{itemize}

\item \textbf{Proposal of the UMMEC Framework}: We propose the UMMEC framework, a novel approach tailored for TFSL scenarios. UMMEC integrates both embedding and classification techniques into a unified framework, addressing critical challenges including intra-class clustering, inter-class separation, and the hubness problem. This comprehensive approach ensures that embeddings are uniformly distributed across the space while maintaining distinct class boundaries, leading to improved performance in TFSL tasks.

\item \textbf{Development of the UMME Embedding Method}: Within the UMMEC framework, the UMME method introduces a decentralized covariance matrix to alleviate hubness and achieve a more uniform embedding distribution. By incorporating an inter-class uniformity loss, UMME ensures that class prototypes are well-spaced in the embedding space, improving class distinction and classification accuracy.

\item \textbf{Design of the UMMC Classification Method}: The UMMC method is a novel classifier that employs a variational Sinkhorn approach to optimize the alignment of embeddings with class centers. By maximizing inter-class distances and minimizing intra-class distances through optimal transport, UMMC iteratively refines class centers, significantly enhancing classification accuracy, especially in data-scarce TFSL scenarios.

\end{itemize}

These innovations collectively advance TFSL by offering a robust framework that improves classification performance with limited labeled data, setting a new benchmark for future research in this field.

\section{Related Work}

\subsection{Few-Shot Learning}

Addressing the challenge of learning from limited data has been a persistent issue in the field. FSL is a specialized domain aimed at enabling models to recognize new classes with only a few examples. Many contemporary approaches to FSL revolve around the meta-learning framework \cite{finn2017model}, which can be broadly classified into three categories: model-based \cite{zhu2024boosting}, optimization-based \cite{zhang2022metanode}, and metric-based methods \cite{zhu2022ease}.

Model-based methods focus on designing specialized networks for FSL tasks. The Local-global Distillation Prototypical Network improves cross-domain few-shot learning by maintaining class affiliation consistency between global and local image crops \cite{zhou2023revisiting}. SAPENet \cite{huang2023sapenet} enhances prototypical networks with multi-head self-attention and channel attention maps, significantly boosting performance on benchmark datasets. Optimization-based methods employ training strategies that facilitate rapid adaptation to new tasks. One notable example is Model-Agnostic Meta-Learning (MAML) \cite{finn2017model}, which seeks an initialization that allows quick fine-tuning with minimal examples. Metric-based methods classify by comparing similarities between examples in an embedding space. Both the support and query sets are mapped into a high-dimensional space using a neural network, and classification is based on proximity. EASE \cite{zhu2022ease} optimizes a linear projection onto a subspace using SVD, and introduces SIAMESE extend clustering techniques to enhance TFSL performance. Furthermore, addressing the hubness problem, \cite{trosten2023hubs} proposes embedding representations on the hypersphere to balance uniformity and local similarity, thereby improving classification accuracy.

\subsection{Transductive Few-Shot Learning}

While inductive FSL relies on a small set of labeled data, TFSL additionally leverages unlabeled test data to improve accuracy under limited labels. However, transductive methods still face significant challenges. For instance, \cite{ma2020transductive} proposed a relation-propagation graph neural network that captures intra-class commonality and inter-class uniqueness, improving performance yet struggling with scalability and complexity on large or complex datasets.

Similarly, \cite{ziko2021transductive} enhanced few-shot learning through constrained clustering with prototype-based objectives and pairwise regularizers. Although effective, this method is sensitive to the initial choice of prototypes, which may lead to suboptimal clustering performance and convergence issues. Furthermore, \cite{zhu2023transductive} advanced clustering and classification by dynamically updating prototype-based graphs. However, its reliance on accurate initial prototypes can result in propagated errors and poor convergence, particularly in complex or imbalanced datasets.

\subsection{Clustering and Classification in Few-Shot Learning}

Clustering and classification are fundamental tasks in the field of FSL, where the goal is to recognize new classes with only a few labeled examples. Clustering-based methods have gained significant attention in FSL due to their ability to group data points into meaningful clusters, which can then be used to infer class labels. 

For instance, the work by \cite{allen2019infinite} proposes infinite mixture prototypes, an adaptive few-shot learning method that improves accuracy and robustness by representing each class with multiple clusters, leading to better capture complex data distributions. Building on these ideas, \cite{ziko2021transductive} introduces a scalable clustering and transductive few-shot learning technique that combines prototype-based objectives with Laplacian regularization, utilizing a concave-convex relaxation and efficient parallelizable optimization, demonstrating that even standard clustering methods can surpass state-of-the-art few-shot learning approaches without complex meta-learning strategies. Additionally, \cite{hu2023adaptive} offers a new clustering method for transductive few-shot classification, utilizing Variational Bayesian inference and Adaptive Dimension Reduction to effectively manage uncertainty and boost accuracy.

To address these persistent challenges in TFSL, we propose the UMMEC Method, which using a decentralized covariance matrix, ensuring a more uniform embedding distribution that enhances robustness and generalization. Second, our method employs adaptive weighting and nonlinear transformation to dynamically balance local similarity and global uniformity, leading to more stable and efficient clustering. Finally, we improve classification accuracy and robustness through a Variational Sinkhorn Few-Shot Classifier, which optimizes distances between samples and class prototypes. These innovations collectively enable UMMEC to outperform existing methods, offering superior performance with minimal labeled data, and advancing the state-of-the-art in TFSL.
\section{Method}

\begin{figure*}[t]
\centering
\includegraphics[width=0.8\textwidth]{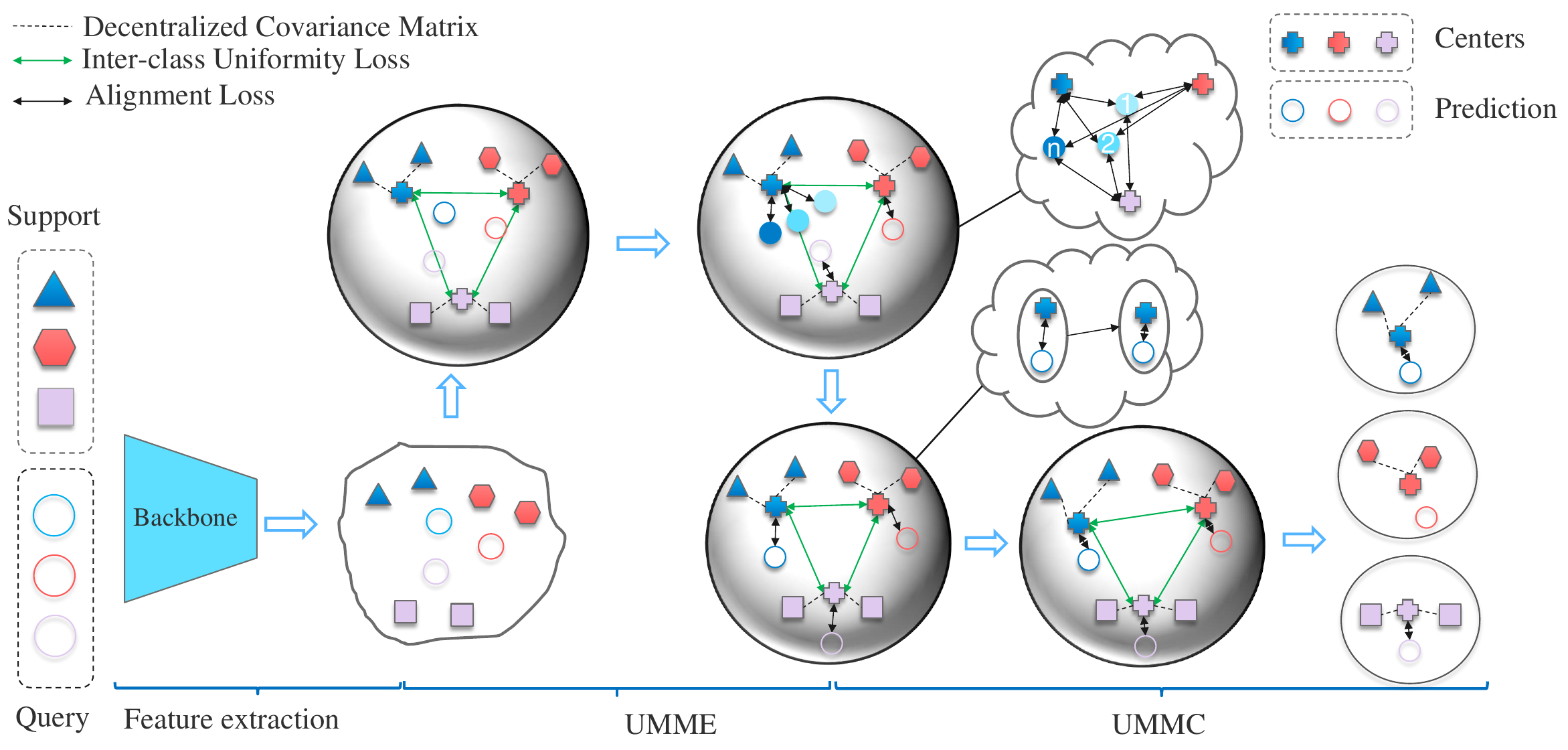} 
\caption{The structure of UMMEC. The UMMEC framework utilizes a pre-trained backbone for feature extraction, applies Decentralized Covariance to achieve inter-class uniformity and intra-class alignment in embeddings, and employs a variational Sinkhorn approach based on optimal transport theory for classifier optimization.}
\label{UMMEC}
\end{figure*}
\subsection{Preliminaries}

The data for the few-shot learning task is divided into three parts: the base set \( \mathcal{D}_{\text{base}} \), the support set \( \mathcal{D}_{\text{support}} \), and the query set \( \mathcal{D}_{\text{query}} \). The base set \( \mathcal{D}_{\text{base}} = \{ ({x}_i, y_i) \mid y_i \in \mathcal{C}_{\text{base}} ;~ i = 1, \ldots , n_{\text{Base}} \} \), contains large-scale labeled samples used for training the feature extractor. \(\mathcal{C}_{\text{base}}\) represents the categories of these samples, providing valuable prior knowledge to describe and understand other samples. In the few-shot scenario, we assume that another labeled dataset \( \mathcal{D}_{\text{novel}} =\{ (x_i, y_i) \mid y_i \in \mathcal{C}_{\text{novel}} ;~ i = 1, \ldots , n_{\text{Novel}} \} \) is given from novel, the unseen classes \(\mathcal{C}_{\text{novel}}\), satisfying \(\mathcal{C}_{\text{base}} \cap \mathcal{C}_{\text{novel}} = \emptyset\). Additionally, we have a test set \(\mathcal{T}_{\text{test}}\) such that \(\mathcal{T}_{\text{test}} \cap \mathcal{D}_{\text{novel}} = \emptyset\), with samples also belonging to \(\mathcal{C}_{\text{novel}}\).

The goal of few-shot learning is to construct a classifier using training samples from both the base set \(\mathcal{D}_{\text{base}}\) and the support set \(\mathcal{D}_{\text{support}}\), capable of accurately classifying the samples in the query set. In an \(N\)-way-\(K\)-shot TFSL problem, tasks are randomly sampled with data from \(N\) novel classes in total and \(K\) examples per class, such that the number of examples in \(S\) is \(L = |S| = N \cdot K\). Following the transductive setting, each task also includes an unlabeled query set \(\mathcal{Q}\) sampled from the same \(N\) classes as the support set \(S\) .

Figure~\ref{UMMEC} provides an overview of our framework. Firstly, we utilize a pre-trained backbone to extract features. Second, we design a unique Decentralized Covariance Matrix paired with inter-class uniformity loss to effectively ensure both inter-class uniformity and class separation. Additionally, we introduce a novel approach that incorporates adaptive weighting and nonlinear transformation into the alignment loss, which preserves intra-class similarity. Finally, we develop an Unbiased Max-Min Classifier method that iteratively updates the class centers and the optimal transport matrix, ensuring that the embeddings are precisely aligned with the learned class centers, thereby maximizing classification accuracy.

\subsection{Unbiased Max-Min Embedding method}
We design an ingenious embedding technique that ensures intra-class clustering while maintaining inter-class separation in a zero-shot learning scenario. We define our embedding function \( E: \mathcal{X} \rightarrow \mathcal{Z} \), where \(\mathcal{X} = \mathbf{x}_{1} \ldots , \mathbf{x}_{n} \in \mathbb{R}^k,n=K(N_S+N_Q)\) is the support and query representations and \(\mathcal{Z} = \mathbf{z}_{1} \ldots , \mathbf{z}_{n} \in \mathbb{S}_d\) is the embedding space.

\subsubsection{Decentralized Covariance Matrix}
\label{sec:decov}

To mitigate \emph{hubness} and promote uniform embeddings, we propose a double-centering operation on the pairwise distance matrix, referred to as the \emph{decentralized covariance matrix}. Let $X$ denote our feature representation (or its reshaped form) such that each column $\mathbf{x}_j$ is in $\mathbb{R}^{m}$. For instance, if a network outputs a $d$-dimensional global-pooled feature, then $m = d$; if it retains spatial dimensions $(h,w)$, then $m = hw$.

\vspace{3pt}\noindent
\textbf{Squared pairwise distances.}
We first construct $\Delta_{e} \in \mathbb{R}^{n\times n}$, where each entry is the squared Euclidean distance between $\mathbf{x}_i$ and $\mathbf{x}_j$:
\begin{equation}
    \Delta_{e}(i,j) 
\;=\; 
\|\mathbf{x}_i - \mathbf{x}_j\|_{2}^{2},
\end{equation}

Expanding the norm yields:
\begin{equation}
    \Delta_e(i,j) 
\;=\;
\|\mathbf{x}_i\|^{2}_{2}
\;+\;
\|\mathbf{x}_j\|^{2}_{2}
\;-\;
2\,\mathbf{x}_i^{\!\top}\mathbf{x}_j,
\end{equation}

Defining $\mathrm{diag}(X^{\!\top}\!X)$ as the vector of squared column norms, we can write $\Delta_{e}$ compactly as
\begin{equation}
\Delta_e 
\;=\; 
\mathrm{diag}(X^{\!\top}X)\,\mathbf{1}^{\!\top}
\;+\;
\mathbf{1}\,\mathrm{diag}(X^{\!\top}X)^{\!\top}
\;-\;
2\,X^{\!\top}X,
\label{eq:delta_e}
\end{equation}
where $\mathbf{1}$ is an all-ones vector of length $n$.

\vspace{3pt}\noindent
\textbf{Euclidean distance matrix.}
By taking an element-wise square root of $\Delta_e$, we obtain
\begin{equation}
\Delta_b 
\;=\;
\sqrt{\Delta_e},
\label{eq:euclid_dist}
\end{equation}
so that $\Delta_b(i,j)$ corresponds to the actual Euclidean distance between $\mathbf{x}_i$ and $\mathbf{x}_j$.

\vspace{3pt}\noindent
\textbf{Double-centering.}
Finally, we \emph{subtract row and column means} (as well as the global mean) from $\Delta_b$ to form the \textbf{decentralized covariance matrix} $\mathcal{D}$:
\begin{equation}
\mathcal{D} 
\;=\;
\Delta_b 
\;-\;
\frac{1}{m}\Bigl(
\mathbf{1}\,\Delta_b
\;+\;
\Delta_b\,\mathbf{1}
\;-\;
\mathbf{1}\,\Delta_b\,\mathbf{1}
\Bigr),
\label{eq:decov_final}
\end{equation}
where $m$ is the dimension factor (e.g.\ $d$ or $hw$), and $\mathbf{1}$ is used appropriately for row or column means. 
Intuitively, this double-centering removes global offsets in $\Delta_b$, preventing certain columns (or samples) from becoming ``universal neighbors'' to many others. Hence, it helps alleviate \emph{hubness} by redistributing pairwise distances more evenly.

\subsubsection{Inter-class Embedding}
\label{sec:interclass}

We next encourage \emph{uniformly spaced} class prototypes while preserving reliable classification. 
Let $S_k$ be the support set of class $k$, containing $K$ labeled samples $\{(\mathbf{z}_j,\,y_j)\mid y_j=k\}$. 
Recalling that $\mathcal{D}(\mathbf{x}_j)$ denotes the decentralized representation of a feature $\mathbf{x}_j$ (see Sec.~\ref{sec:decov}), we define each class prototype $\mathbf{P}_k$ by:
\begin{equation}
\label{eq:Pk}
\mathbf{P}_k 
= 
\frac{1}{K}\,\sum_{(\mathbf{z}_j,\,y_j)\,\in\,S_k} 
\mathcal{D}(\mathbf{x}_j),
\end{equation}
Here, $\mathbf{x}_j$ could be a $d$-dimensional embedding (after global pooling) or a reshaped feature map.

\vspace{3pt}\noindent
\textbf{Uniformity loss.}
To promote \emph{uniform} separation among these prototypes, we introduce:
\begin{equation}
    \mathcal{L}_{\text{uniformity}} = \frac{1}{N(N-1)} \sum_{c=1}^N \sum_{\substack{j=1 \\ j \neq c}}^N \exp \left( -\frac{\|\mathbf{P}_c - \mathbf{P}_j\|_F^2}{\gamma^2} \right),
\end{equation}
where $\|\cdot\|_F$ is the Frobenius norm. Because the exponential term becomes large if two prototypes are too close, minimizing $\mathcal{L}_{\text{uniformity}}$ naturally \emph{pushes} distinct $\mathbf{P}_c$ and $\mathbf{P}_j$ farther apart, thus encouraging a more uniform arrangement of class prototypes.

\vspace{3pt}\noindent
\textbf{Classification loss.}
We also define a classification loss on a query set $Q=\{\bigl(\mathbf{e}_i,y_i\bigr)\}$ sampled from the same $N$ classes. 
While $Q$ is unlabeled at test time, in \emph{meta-training episodes} we do have access to $y_i$ for each query sample, enabling us to compute:
\begin{equation}
    \mathcal{L}_{\text{classification}} = - \frac{1}{|Q|} \sum_{(\mathbf{e}_i, y_i) \in Q} \log \frac{\exp \left( -\|\mathbf{S}_i - \mathbf{P}_{y_i}\|_F^2 \right)}{\sum_{c=1}^N \exp \left( -\|\mathbf{S}_i - \mathbf{P}_c\|_F^2 \right),}
\end{equation}
where $\mathbf{S}_i = \mathcal{D}(\mathbf{e}_i)$ is the decentralized representation of the query feature. 
Intuitively, if $\mathbf{S}_i$ is closest to $\mathbf{P}_{y_i}$, the numerator in the softmax dominates, reducing the loss.

\vspace{3pt}\noindent
\textbf{Combined objective.}
We combine these two losses into:
\begin{equation}
\label{eq:global}
\mathcal{L}_{\text{global}}
=
\alpha\,\mathcal{L}_{\text{uniformity}}
+\, 
(1-\alpha)\,\mathcal{L}_{\text{classification}},
\end{equation}
where $\alpha\in(0,1)$ balances the emphasis on \emph{prototype spread} vs.\ \emph{classification accuracy}. 
Minimizing $\mathcal{L}_{\text{global}}$ thus enforces both uniform prototype placement and robust per-query classification in few-shot learning tasks.

\subsubsection{Intra-class Embedding (Revised)}
\label{sec:intra_class}

While our global loss (Sec.~\ref{sec:interclass}) encourages clear separation among different classes, 
we also need to preserve \emph{local similarity} structures within each class. 
To this end, we propose a local alignment loss $\mathcal{L}_{\text{local}}$ 
incorporating \emph{adaptive weighting} and a \emph{nonlinear transformation}.

\vspace{3pt}\noindent
\textbf{Adaptive weighting.}
Let $p_{ij}$ be the pairwise similarity between two embeddings $\mathbf{z}_i$ and $\mathbf{z}_j$. 
Throughout this subsection, $\{\mathbf{z}_i\}$ are \emph{original feature embeddings} 
, \emph{not} the decentralized covariance $\mathcal{D}(\cdot)$ used in Sec.~\ref{sec:decov}; 
the local alignment acts directly on $\mathbf{z}$ space to better capture immediate neighborhood structure. 
We define 
\begin{equation}
\label{eq:beta_adaptive}
\beta_{ij} \;=\; 
\frac{\exp\!\bigl(\lambda \,p_{ij}\bigr)}{\sum_{k}\exp\!\bigl(\lambda \,p_{ik}\bigr)},
\end{equation}
where $\lambda$ is a temperature parameter controlling the weight distribution's sharpness. 
High-similarity pairs $(i,j)$ yield larger $\exp(\lambda p_{ij})$ and thus higher $\beta_{ij}$, so the model pays more attention to pairs that are already quite similar. 
\emph{Note} that the similarity $p_{ij}$ may be computed for (i) all pairs in a mini-batch (support + query), 
or (ii) pairs \emph{only} within the same class if we wish to focus on intra-class neighbors. 
In our implementation, we primarily consider \emph{all} pairs $(\mathbf{z}_i,\mathbf{z}_j)$ in the current episode, but one can restrict to same-class pairs depending on the task.

\vspace{3pt}\noindent
\textbf{Nonlinear transformation.}
Next, to model complex relationships among embeddings, 
we apply a parameterized Sigmoid $\psi(\cdot)$ to the \emph{cosine similarity} $\mathbf{z}_i^\top \mathbf{z}_j$:
\begin{equation}
\label{eq:psi_function}
\psi\bigl(\mathbf{z}_i^\top \mathbf{z}_j\bigr)
\,=\,
\sigma\!\bigl(\mu\,\mathbf{z}_i^\top \mathbf{z}_j\bigr),
\end{equation}
where $\mu$ is a learnable parameter, and $\sigma(\cdot)$ is the standard Sigmoid function. 
Since $\mathbf{z}_i^\top \mathbf{z}_j$ typically lies in $[-1,1]$ when $\mathbf{z}_i,\mathbf{z}_j$ are $\ell_2$-normalized, 
scaling by $\mu$ allows the model to adjust how steeply $\psi$ changes with respect to that similarity measure. 

\vspace{3pt}\noindent
\textbf{Local alignment loss.}
Combining these ideas, our local alignment loss is:
\begin{equation}
\label{eq:local_loss}
\mathcal{L}_{\text{local}}
\,=\;
-\,
\sum_{i,j}
\beta_{ij}\,
p_{ij}\,
\psi\bigl(\mathbf{z}_i^\top \mathbf{z}_j\bigr),
\end{equation}
Here, $p_{ij}$ represents the \emph{original similarity} used to define the adaptive weights in \eqref{eq:beta_adaptive}. 
Multiplying $p_{ij}$ and $\psi(\mathbf{z}_i^\top \mathbf{z}_j)$ is not redundant: $p_{ij}$ essentially serves as a \emph{reference similarity} 
for weighting, while $\psi(\mathbf{z}_i^\top \mathbf{z}_j)$ adds a \emph{nonlinear scaling} that can modulate how strongly each pair influences the local alignment. 

\vspace{3pt}\noindent
\textbf{Interpretation and example.}
If $\mathbf{z}_i$ and $\mathbf{z}_j$ belong to the same class, $p_{ij}$ and $\mathbf{z}_i^\top \mathbf{z}_j$ should both be relatively high, 
so their product is strongly weighted by $\beta_{ij}$. 
This \emph{does not} imply we want to reduce such similarity; rather, we further amplify it to ensure these points remain close in the embedding space. 
By contrast, if two samples have lower similarity, they receive less weight, 
and the model invests fewer resources trying to align their features. 

\subsubsection{Final Combined Loss}
We combine the local alignment loss $\mathcal{L}_{\text{local}}$ with the global loss $\mathcal{L}_{\text{global}}$ by
\begin{equation}
\label{eq:final_combined_loss}
\mathcal{L}_{\text{total}}
\,=\,
\eta\,\mathcal{L}_{\text{local}}
\;+\;
(1 - \eta)\,\mathcal{L}_{\text{global}},
\end{equation}
where $\eta\in(0,1)$ controls the trade-off between \emph{intra-class refinement} and \emph{inter-class separation}. 
Minimizing $\mathcal{L}_{\text{total}}$ thus balances fine-grained local consistency with global class separation. 

\subsection{Unbiased Max-Min Classifier Method}
\label{sec:ummc}

We now refine our embeddings $\{ \mathbf{z}_1, \ldots, \mathbf{z}_n\}$ by iteratively updating \emph{class centers} via an \textbf{Optimal Transport} (OT) perspective. 
This process is realized by the \textbf{Variational Sinkhorn Few-Shot Classifier}, which maximizes inter-class separation while minimizing intra-class distances.

\vspace{3pt}
\noindent
\textbf{Source \& Target Distributions.}
In the OT framework, we interpret:
\begin{itemize}[leftmargin=1.2em,topsep=3pt,itemsep=1pt]
\item \emph{Source} distribution as the set of \emph{all} embedded samples, each represented by $\mathbf{z}_i$ and collectively forming an empirical distribution over $n$ points.
\item \emph{Target} distribution as $K$ classes, each associated with a certain mass that reflects how many samples should ideally belong to that class. For an $N$-way $K$-shot setup, typically $K = N$, or we might fix $K=N$ and let each class have equal mass if uniform class sizes are assumed. 
\end{itemize}
Our goal is to find a \emph{transport plan} $P$ that aligns each sample $\mathbf{z}_i$ (from source) to one of $K$ class centers, subject to the correct marginal constraints (i.e.\ the sum of $P_{ik}$ over $i$ matches the target mass for class $k$).

\vspace{3pt}
\noindent
\textbf{Initialization of Class Centers.}
We begin by computing an \emph{initial} center $\widetilde{c}_k$ for class $k$ from support samples:
\begin{equation}
\label{eq:init_center}
\widetilde{c}_k
=
\frac{1}{\lvert S_k\rvert}
\sum_{(\mathbf{z}_i,\, y_i)\,\in\,S_k}
\mathbf{z}_i,
\end{equation}
where $S_k$ gathers the support points of class $k$. 
Let $M\in \mathbb{R}^{n\times K}$ be the \emph{cost matrix} whose element $M_{ik} = \| \mathbf{z}_i - \widetilde{c}_k \|_{2}^{2}$. 
Our objective is two-fold: (i) \emph{minimize} intra-class distances (so points are near their assigned center) and (ii) \emph{maximize} inter-class distances (so centers are well separated). 

\vspace{3pt}
\noindent
\textbf{Variational Sinkhorn Method.}
We introduce regularization via the Kullback--Leibler term $D_{\mathrm{KL}}(P\|r\,c^\top)$, 
leading to the objective:
\begin{equation}
\label{eq:OT_objective}
\min_{P\in \mathcal{U}(r,c)}
\;\bigl\langle P,\,M\bigr\rangle
\;+\;
\frac{1}{\lambda}\,D_{\mathrm{KL}}(P\;\|\;r\,c^\top).
\end{equation}
Here, $\mathcal{U}(r,c)\! =\!\{\, P\ge 0 \mid P\,\mathbf{1}=r,\, P^\top\mathbf{1}=c\}$ enforces marginals $r,c$. 
Algorithm~\ref{alg:algorithm} (a \emph{Sinkhorn} procedure) iteratively updates \emph{scaling vectors} $u,v$ until $P$ meets those marginals, 
then refines each class center based on the newly adjusted transport plan.

\begin{algorithm}[tb]
\caption{Variational Sinkhorn Few-Shot Classifier (Sketch)}
\label{alg:algorithm}
\textbf{Input:} Embeddings $Z = \{\mathbf{z}_1,\ldots,\mathbf{z}_n\}$, initial centers $\widetilde{C}=\{\widetilde{c}_1,\ldots,\widetilde{c}_K\}$, max iterations $T$ \\
\textbf{Parameter:} regularization $\lambda$, step size $\alpha$ \\
\textbf{Output:} updated centers $\widetilde{C}$, transport matrix $P$
\begin{algorithmic}[1]
\STATE Initialize scaling vectors $u=\mathbf{1}_n,\;v=\mathbf{1}_K$
\FOR{$t=1$ to $T$}
  \STATE \textit{// Update scaling $u,v$ for Sinkhorn}
  \WHILE{not converged}
    \STATE $u \gets r / (P\,v)$
    \STATE $v \gets c / (P^\top u)$
    \STATE $P \gets \mathrm{diag}(u)\,P\,\mathrm{diag}(v)$
  \ENDWHILE
  \STATE \textit{// Update class centers}
  \STATE $\Omega \gets$ 
         \ $\Omega = \tfrac{P^\top Z}{\sum_i P^\top_{i,\cdot}}$
  \STATE $\widetilde{c}_k \gets \widetilde{c}_k + \alpha\,(\Omega_{k} - \widetilde{c}_k),\; \forall k$
\ENDFOR
\RETURN $\widetilde{C},\,P$
\end{algorithmic}
\end{algorithm}

\vspace{3pt}
\noindent

\vspace{3pt}
\noindent
\textbf{Comparison to Initial Centers.}
Updating $\widetilde{C}$ with the transport plan $P$ typically yields significantly better class separation than the raw averages in \eqref{eq:init_center}. 
We find that the final centers (after $T$ iterations) are \emph{less sensitive} to outliers and more discriminative. 
Section~X.X provides quantitative evidence that accuracy can improve by several percentage points compared to using the naive prototype average.

\vspace{3pt}
\noindent
\textbf{Final Classification of Queries.}
After convergence, we obtain:
\begin{itemize}[leftmargin=1.2em,itemsep=0pt,topsep=3pt]
\item updated class centers $\widetilde{C} = \{\widetilde{c}_1,\dots,\widetilde{c}_K\}$,
\item optimal transport matrix $P$ that assigns each sample $\mathbf{z}_i$ to classes.
\end{itemize}
To classify a new query $\mathbf{z}_q$, we compute its squared distance to each $\widetilde{c}_k$, and assign:
\begin{equation}
\label{eq:query_classification}
\widehat{y}_q 
\;=\;
\arg\min_{k}\;
\|\mathbf{z}_q - \widetilde{c}_k\|_{2}^{2}.
\end{equation}
Alternatively, one can use the learned $P$ if $\mathbf{z}_q$ was included in the transport step. Both approaches converge to similar decisions once the centers are updated.

\section{Experiments}
In this section, we present the experimental evaluation of our method. We start by outlining the experimental settings. Subsequently, we conduct ablation studies to analyze the individual contributions of various components in our approach. Finally, we compare the performance of our method against other state-of-the-art (SOTA) techniques.

\begin{table*}[t]
\centering
\caption{Performance comparison of different methods on Mini-ImageNet, Tiered-ImageNet, and CUB datasets under 1-shot and 5-shot settings. THE VALUES IN BOLD ARE THE BEST IN EACH BACKBONE.}
\label{acc}
\scalebox{0.9}{
\begin{tabular}{*{9}{c}}
\toprule
\textbf{Method} & \textbf{Setting} &\textbf{Backbone} & \multicolumn{2}{c}{\textbf{Mini-ImageNet}} & \multicolumn{2}{c}{\textbf{Tiered-ImageNet}} & \multicolumn{2}{c}{\textbf{CUB}} \\
\cmidrule(lr){4-5}\cmidrule(lr){6-7}\cmidrule(lr){8-9}
 & & & \textbf{K=1} & \textbf{K=5} & \textbf{K=1} & \textbf{K=5} & \textbf{K=1} & \textbf{K=5} \\
\midrule
TIM\cite{boudiaf2020information} & Transductive & ResNet-18 & 67.30 & 79.80 & 74.10 & 84.10 & 82.87 & 91.58 \\
$\alpha$-TIM\cite{veilleux2021realistic}& Transductive & ResNet-18 & 67.40& 82.50 & 74.40 & 86.60 &75.70& 89.80  \\ 
SLK-MS\cite{ziko2021transductive} & Transductive & ResNet-18 & 73.10 & 82.82 & 79.99 & 86.55  & 81.88 & 88.55 \\ 
BAVARDAGE\cite{hu2023adaptive} & Transductive & ResNet-18 & 75.10 & 81.50 & 80.30 & 87.10 & \textbf{87.40} & \textbf{92.00 }\\ 
UMMEC (Ours) & Transductive & ResNet-18 & \textbf{77.35} & \textbf{83.93} & \textbf{82.67} & \textbf{88.26} & 86.91 & 91.35 \\ 
\midrule
SimpleShot\cite{wang2019simpleshot} & Inductive & WRN-28-10 & 63.50 & 80.33 & 69.75 & 85.31 & 82.80 & 86.30 \\ 
RankDNN\cite{guo2023rankdnn}  & Inductive & WRN-28-10 &66.67& 84.79& 74.00& 88.80& 81.78 & 91.12\\ 
EPNet\cite{rodriguez2020embedding}& Transductive & WRN-28-10 & 70.74&  84.34& 78.50& 88.36&87.75 & 94.03 \\ 
LaplacianShot\cite{ziko2020laplacian}& Transductive & WRN-28-10&74.86 & 84.13 & 80.18 & 87.56 & 80.18 & 87.56\\ 
EASE+SIAMESE\cite{zhu2022ease} & Transductive & WRN-28-10 & 81.19 & 87.82 & 82.04 & 88.06 & 91.99 & 94.36 \\ 
NOHUB-S+SIAMESE\cite{trosten2023hubs} & Transductive & WRN-28-10 & 82.00 & 88.03 & 82.85 & 88.31 & 92.63 & 94.69 \\ 
\(\text{AM}_\text{PLC}\)\cite{lazarou2024adaptive} & Transductive & WRN-28-10 & 80.99 & 87.86 & 85.26 & 90.30 & 91.32 & 94.14 \\ 
UMMEC (Ours) & Transductive & WRN-28-10 & \textbf{82.55} & \textbf{88.64} & \textbf{87.83} & \textbf{91.82} & \textbf{93.65} & \textbf{95.56} \\
\bottomrule
\end{tabular}}
\end{table*}

\subsection{Experimental Settings}
\subsubsection{Datasets}
In this paper, we evaluate our method on three benchmark datasets: Mini-ImageNet \cite{vinyals2016matching}, Tiered-ImageNet \cite{ren2018meta}, and CUB-200-2011 \cite{welinder2010caltech}.

\textbf{Mini-ImageNet} consists of 100 categories, each containing 600 images. It is divided into three parts: 64 base categories for training, 16 novel categories for validation, and the remaining 20 categories for testing. 

\textbf{Tiered-ImageNet} contains 779,165 images from 608 categories, where 351 base categories are used for training, 97 novel categories for validation, and the remaining 160 novel categories for testing. 

\textbf{CUB-200-2011} dataset, known for its fine-grained bird species classification, consists of 200 bird species with a total of 11,788 images. 

These datasets provide a comprehensive evaluation framework for assessing the performance and generalization capability of our method across diverse image categories and levels of granularity.

\subsubsection{Evaluation}

We adhere to the standard evaluation protocol in FSL and measure accuracy for both 1-shot and 5-shot classification tasks, using 15 images per class in the query set. Our evaluation is conducted over 10,000 episodes, which is the standard practice in FSL.

\subsubsection{Implementation details}
Our implementation is based on PyTorch. We optimize our methods for 150 iterations using the Adam optimizer \cite{kingma2014adam} with a learning rate of \( \eta = 0.1 \).

\subsubsection{Backbones}
To ensure fair comparison, we follow the practice of most TFSL studies by using the open-source ResNet-18 and WideResNet-28-10 (WRN-28-10) backbones for feature extraction, as provided in the works of \cite{veilleux2021realistic} and \cite{mangla2020charting}.

\subsection{Results}
\subsubsection{Comparisons with Other Methods}
To demonstrate the effectiveness of our approach in addressing the challenges of FSL, we evaluated its performance against state-of-the-art FSL methods on the Mini-ImageNet, Tiered-ImageNet, and CUB datasets. We assessed the performance of all datasets using our method under both 5-way-1-shot and 5-way-5-shot settings, comparing it with leading inductive\cite{wang2019simpleshot, guo2023rankdnn} and transductive\cite{boudiaf2020information,rodriguez2020embedding,ziko2020laplacian,veilleux2021realistic, zhu2022ease, trosten2023hubs, lazarou2024adaptive} FSL approaches, including several clustering-based FSL methods \cite{ziko2021transductive, hu2023adaptive}.

As shown in the table \ref{acc}, our method outperforms all baseline methods across both settings for all datasets. 
In the ResNet-18 backbone setting, the proposed UMMEC method achieves the highest accuracy across all datasets and settings, demonstrating superior performance in both 1-shot and 5-shot scenarios. Specifically, UMMEC outperforms clustering methods, BAVARDAGE and SLK-MS, with notable improvements in the Mini-ImageNet and Tiered-ImageNet datasets.

For the WRN-28-10 backbone setting, UMMEC also shows the best performance, surpassing other advanced methods like NOHUB-S+SIAMESE and \(\text{AM}_\text{PLC}\). The significant of UMMEC gains, especially on the Tiered-ImageNet and CUB datasets, highlight its effectiveness in handling complex few-shot learning tasks. This indicates that our approach provides valuable FSL embeddings and classification, setting a new benchmark for the best TFSL performance.

\begin{figure*}[t]
\centering
\includegraphics[width=0.8\textwidth]{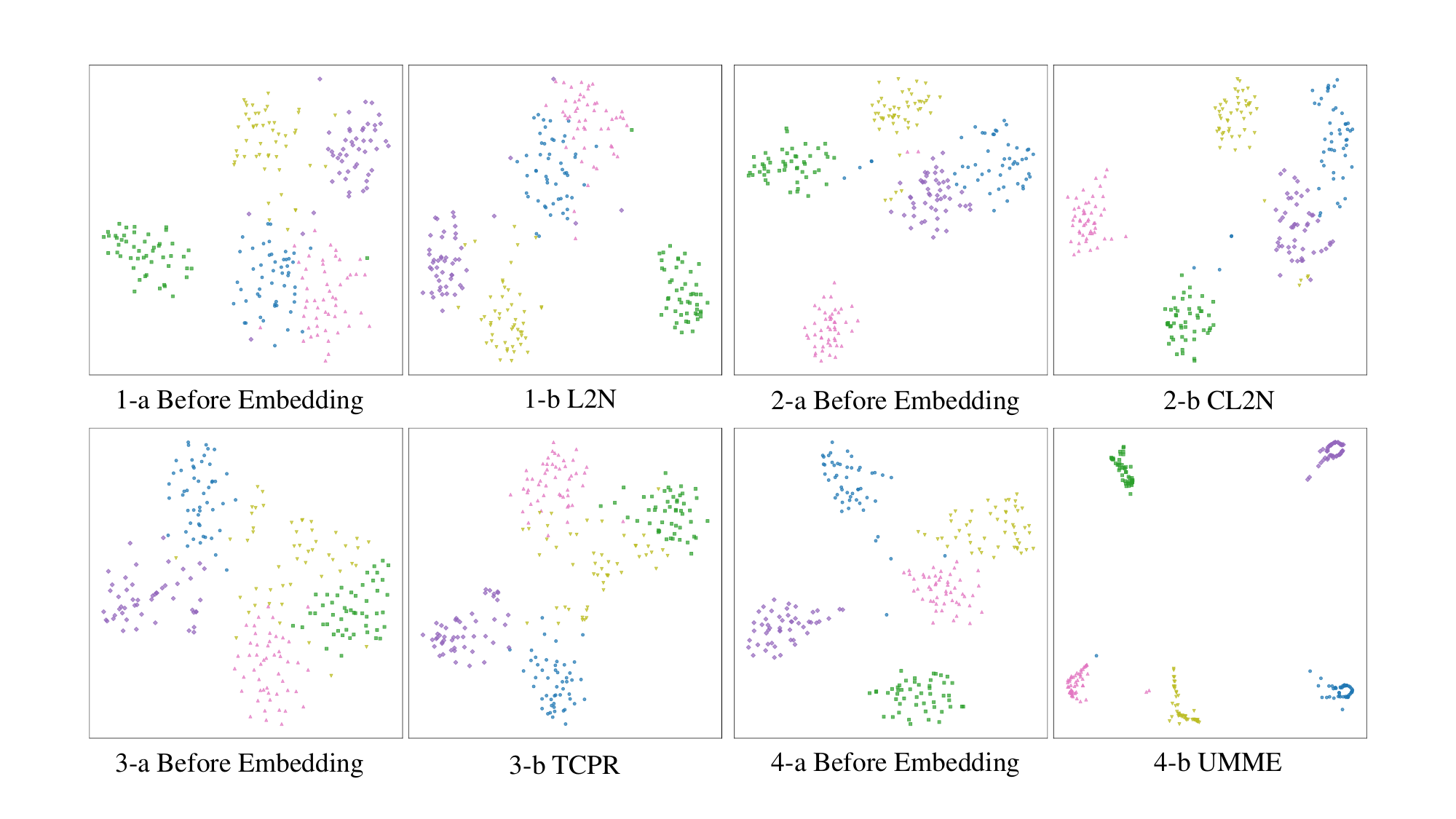} 
\caption{The results of the t-SNE experiment are shown in the figure. Figure (a) depicts the class distribution before embedding, while Figure (b) shows the class distribution after applying the L2N\cite{wang2019simpleshot}, CL2N\cite{wang2019simpleshot}, TCPR\cite{xu2022alleviating} and our UMME methods. The L2N, CL2N, TCPR methods exhibit some degree of class confusion, with indistinct class boundaries. In contrast, our UMME method achieves clear inter-class separation and strong intra-class clustering, resulting in well-defined class boundaries.}
\label{tsne}
\end{figure*}

\subsection{Ablation Study}
In the ablation experiments of this paper, we analyze the effects of the loss functions \(\mathcal{L}_{\text{local}}\) and \(\mathcal{L}_{\text{global}}\) as well as the UMMC. We used the Tiered-ImageNet dataset and employed WideResNet-28-10 as the feature extractor. 

\begin{table}[t]
\centering
\caption{Ablation experiment results (\%) on the Tiered-ImageNet with the backbone of WideResNet-28-10. THE VALUES IN BOLD ARE THE BEST.}
\label{Ablation}
\begin{tabular}{*{5}{c}}
\toprule
\(\mathcal{L}_{\text{local}}\) & \(\mathcal{L}_{\text{global}}\) & UMMC & \textbf{K=1} & \textbf{K=5} \\
\midrule
\ding{51} & \ding{51} & \ding{55} & 72.74 & 76.93 \\
\ding{51} & \ding{55} & \ding{51} & 65.71 & 79.13 \\
\ding{55} & \ding{51} & \ding{51} & 75.29 & 84.88 \\
\ding{51} & \ding{51} & \ding{51} & \textbf{87.83} & \textbf{91.82} \\
\bottomrule
\end{tabular}
\end{table}

The results presented in Table~\ref{Ablation} clearly demonstrate the importance of each component in our model. When the UMMC module is excluded and a simple classifier is used instead, the performance significantly decreases at \(K=1\), although it remains relatively stable at \(K=5\). This indicates that while the UMMC module is crucial for improving the model’s performance in 1-shot scenarios, it may be less critical when more examples are available for learning.

The omission of the \(\mathcal{L}_{\text{local}}\) loss function also leads to a decrease in performance, particularly at \(K=5\), which suggests that this loss plays a vital role in preserving local similarity across multiple examples. Its contribution is essential for maintaining high accuracy when the model is exposed to several examples per class, as seen by the significant drop in 5-shot accuracy when this loss is removed.

Interestingly, the \(\mathcal{L}_{\text{global}}\) loss function appears to be more critical in 1-shot scenarios. The model achieves a higher \(K=1\) accuracy when \(\mathcal{L}_{\text{global}}\) is included, even if \(\mathcal{L}_{\text{local}}\) or UMMC are excluded. This indicates that the \(\mathcal{L}_{\text{global}}\) loss is essential for capturing global patterns in situations with limited data.

Finally, when all components are used together, the model achieves the best performance at both \(K=1\) and \(K=5\), highlighting the complementary roles each plays in enhancing the model’s few-shot learning capabilities.

\subsection{Visualization Analysis of The Results}

We applied the UMME embedding method to the CUB dataset, leveraging a WRN backbone, and evaluated it under a 5-way setting. The t-SNE visualization, illustrated in Figure~\ref{tsne}, highlights the performance of our embedding approach. The t-SNE plots reveal that our embedding method excels in both separating different classes and ensuring strong intra-class cohesion. This dual capability not only leads to well-defined and distinct class boundaries but also enhances the overall reliability and effectiveness of the model, particularly in few-shot learning.

In addition to these observations, we conducted a comparative analysis with other widely-used embedding techniques such as L2N\cite{wang2019simpleshot}, CL2N\cite{wang2019simpleshot}, and TCPR\cite{xu2022alleviating}. The results clearly demonstrate that, while these methods struggle with class confusion and indistinct boundaries, our UMME embedding method maintains a clear separation between classes and achieves superior intra-class clustering. This distinct advantage underscores the robustness and precision of our approach, positioning it as a superior embedding solution within the domain of few-shot learning.

\section{Conclusion}

In conclusion, the UMMEC Method represents a significant advancement in the realm of few-shot learning, addressing the pressing challenges posed by the scarcity of annotated data. By introducing a decentralized covariance matrix, UMMEC effectively mitigates the hubness problem, ensuring a more balanced and uniform embedding distribution. The integration of combined local alignment and global uniformity through adaptive weighting and nonlinear transformation ensures that intra-class similarities are preserved while maintaining clear inter-class distinctions. Furthermore, the incorporation of the Variational Sinkhorn Few-Shot Classifier optimizes the relationships between samples and class prototypes, enhancing both accuracy and robustness in classification tasks. Through these innovations, UMMEC not only overcomes the limitations of existing methods but also sets a new benchmark in few-shot learning performance. This work opens avenues for exploring more sophisticated models that can further bridge the gap between limited data scenarios and high-accuracy machine learning applications. 
{
    \small
    \bibliographystyle{ieeenat_fullname}
    \bibliography{main}
}

\end{document}